\documentclass{article}
\usepackage{graphicx} 
\usepackage[square,sort,comma,numbers]{natbib}
\usepackage[preprint]{neurips_2024}
\usepackage[utf8]{inputenc} 
\usepackage[T1]{fontenc}    
\usepackage{hyperref}       
\usepackage{url}            
\usepackage{booktabs}       
\usepackage{amsfonts}       
\usepackage{nicefrac}       
\usepackage{microtype}      
\usepackage[table,x11names]{xcolor}
\usepackage{multirow}       
\usepackage{float}          
\usepackage{appendix}       

\usepackage[hang,flushmargin]{footmisc}
\usepackage{pdfpages}
\usepackage{xcolor}
\definecolor{lightergray}{gray}{0.9}

\usepackage{array}

\usepackage{pgfplots}
\pgfplotsset{width=10cm,compat=1.9}

\usepackage{tikz}

\begin{document}

\title{1.5-Pints Technical Report: \\ Pretraining in Days, Not Months -- Your Language Model Thrives on Quality Data}
\author{Calvin Tan
$\quad$ Jerome Wang\\
\\
Pints.ai Labs \\
\texttt{\{calvin, jerome\}@pints.co}\\
}

\maketitle

\begin{abstract}
    This paper presents a compute-efficient approach to pre-training a Language Model -- the "1.5-Pints" -- in only 9 days, while outperforming state-of-the-art models as an instruction-following assistant. Based on MT-Bench (a benchmark that emulates human judgments), 1.5-Pints outperforms Apple’s OpenELM and Microsoft’s Phi. This is achieved by a carefully curated pre-training dataset of 57 billion tokens, using a mix of automated workflows and manual human review. The selection of the dataset prioritizes content that is considered expository and "textbook-like" to aid the model in reasoning and logical deduction, culminating in its overall ability as a strong and versatile AI model. In terms of the model architecture, we employed a modified Mistral tokenizer, alongside a Llama-2 architecture for wider compatibility. For training, we adopted the methodologies used by StableLM, TinyLlama, and Huggingface Zephyr. 1.5-Pints demonstrates that by focusing on data quality over quantity in LLM training, we can significantly reduce training time and resources required. We believe this approach will not only make pre-training more accessible but also reduce our carbon footprint. Our findings and resources\footnote{Code can be found at \href{https://github.com/Pints-AI/1.5-Pints}{https://github.com/Pints-AI/1.5-Pints} and Dataset can be found at \href{https://huggingface.co/datasets/pints-ai/Expository-Prose-V1}{https://huggingface.co/datasets/pints-ai/Expository-Prose-V1}} from this research are open-sourced, aiming to facilitate further advancements in the field. The 1.5-Pints model is available in two versions: 2K\footnote{\href{https://huggingface.co/pints-ai/1.5-Pints-2K-v0.1}{https://huggingface.co/pints-ai/1.5-Pints-2K-v0.1}} and 16K\footnote{\href{https://huggingface.co/pints-ai/1.5-Pints-16K-v0.1}{https://huggingface.co/pints-ai/1.5-Pints-16K-v0.1}} context windows.

\begin{table}[H]
  \caption{MT-Bench Comparison with SOTA Compact Models}
  \label{MT-Bench-results}
  \centering
  \begin{tabular}{lccc}
    \toprule
    Model & Score & Pre-train Tokens & Parameter Size\\
    \midrule
    1.5-Pints-2K & \textbf{3.73} & \textbf{0.115T} & 1.57B\\
    apple/OpenELM-1\_1B-Instruct & 3.34 & 1.8T & 1B\\
    microsoft/phi-1\_5 & 3.33 & 0.15T & 1.3B\\
    databricks/dolly-v2-3b & 2.33 & 0.3T & 3B\\
    EleutherAI/pythia-2.8b & 1.81 & 0.3T & 2.8B\\
    tiiuae/falcon-rw-1b & 1.18 & 0.35T & 1B\\
    \bottomrule
  \end{tabular}
\end{table}
    
\end{abstract}

\section{Introduction}
In recent years, Large Language Models (LLMs) have arguably been the pinnacle of language modeling, possessing excellent language, reasoning, and logical deduction capabilities. However, training LLMs has been a resource-intensive endeavor. For example, Meta relies on two of its custom-built 24K H100 GPU clusters to train the Llama-3 family~\cite{meta-llama-3}. The bulk of GPU resource requirements stem from the attention mechanism within the transformer architecture~\cite{vaswani2023attention}, which scales quadratically with input sequence (or context length)~\cite{ren2021combiner}, resulting in a hefty amount of computation. 

To reduce the infrastructure and resources required to run transformer models, many techniques focus on hardware optimizations~\cite{abts2022tensor} and hardware-aware algorithms~\cite{DBLP:journals/corr/abs-2109-03389, dao2023flashattention2}, attention approximations~\cite{NEURIPS2023_6f9806a5,9065498}, or selective attention mechanisms~\cite{DBLP:journals/corr/abs-2403-01590, pmlr-v162-hua22a,ren2021combiner}. 

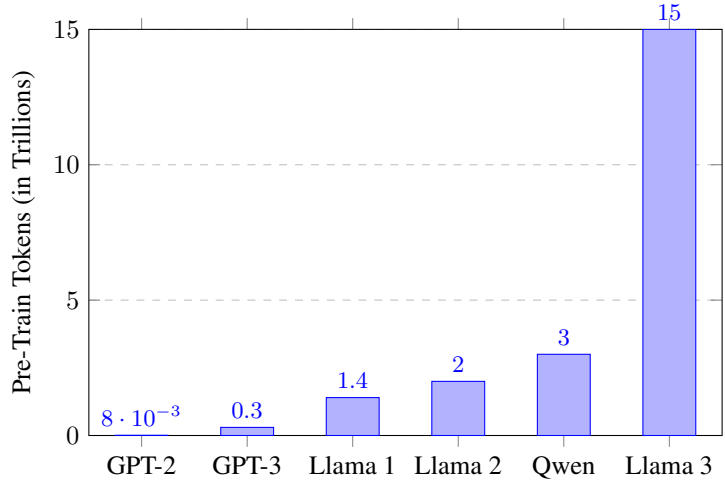
\begin{figure}[h]
    \centering
    \caption{Growth in Pre-Training Corpus}
    \label{figure:growth-pretraining-corpus}
    \begin{tikzpicture}
    \begin{axis}[
        height=0.5\textwidth,
        ybar,
        symbolic x coords={GPT-2, GPT-3, Llama 1, Llama 2, Qwen, Llama 3},
        xtick=data,
        ylabel={Pre-Train Tokens (in Trillions)},
        ytick={0,5,10,15},
        ymin=0, ymax=15,
        ymajorgrids=true,
        grid style=dashed,
        bar width=20pt,
        nodes near coords,
        nodes near coords align={vertical},
        every node near coord/.append style={font=\small},
    ]
    
    \addplot
        coordinates {
            (GPT-2,0.008)(GPT-3,0.3)(Llama 1,1.4)(Llama 2,2)(Qwen,3)(Llama 3,15)
        }; 
    \end{axis}
    \end{tikzpicture}
\end{figure}

Reduction of training corpus is also another way. This can be achieved by improving the quality of the training corpus, as it is well-established that better data leads to better models~\cite{maxmarion2023when, yang2023dataset, xia2024less}. However, the growth in the size of the training corpus continues to trend upwards indefinitely (see figure~\ref{figure:growth-pretraining-corpus}), which makes quality control increasingly difficult.

Anecdotally, this is akin to making a student spend more time reading a larger set of materials to figure out what is relevant. Therefore, although improving data quality for training LLMs is not a novel idea, there is still a lack of more sophisticated and effective methods to increase data quality. Conversely, a meticulously crafted syllabus would help a student learn in a shorter timeframe. Datasets could similarly be meticulously crafted to optimize LLM learning.

We posit that this is an important research area, as a breakthrough will significantly reduce training time and resources required. This can spur greater research advances, proliferate useful and desirable AI products, while reducing environmental impact. The feasibility of some aspects of training corpus refinement has been demonstrated in TinyStories~\cite{eldan2023tinystories} and Phi~\cite{gunasekar2023textbooks, li2023textbooks}, where models trained with synthetically generated data (using GPT3.5 and GPT4) achieved remarkable performance. However, Phi's training corpus was not released, limiting reproducibility and further research. Moreover, a dataset generated by GPT3.5 or GPT4 would come with restrictions that limit its free use and commercial applications. Accordingly, we aim to replicate this approach and fully release our findings, code, and resources, encouraging the research community to build upon our work.

\section{Our Contributions}
We constructed a pre-training corpus of 57 billion tokens, comprising mainly expository prose collected from high-quality sources such as research papers, copyright-free books, parliamentary debates, and synthetically generated content. See Section~\ref{section:pre-training-dataset} for a detailed breakdown.

Using our pre-training corpus, we trained a 1.56 billion parameter model, which we named 1.5-Pints (pronounced "one-and-a-half" Pints). The training took a total of 9 days on 8 A100s, with a total of 115 billion tokens across pre-training, fine-tuning, and direct preference optimization. For evaluation, we used MT-Bench, a benchmark that substitutes GPT-4 for human evaluators to assess model responses, thereby overcoming limitations of traditional benchmarks in measuring adherence to instructions and usefulness~\cite{zheng2023judging}. 1.5-Pints outperforms its peers on MT-Bench, despite using 15 to 25 times less training data. Of note, 1.5-Pints is also available in 16K context window version, which is twice that of Llama-3, and is generally 4-8 times larger than other notable models. This allows 1.5-Pints to handle a wider range of tasks, such as large passage summarization and extended multi-turn dialogues.

\section{Approach}
\subsection{Creating the pre-training dataset}

We believe that an optimal LLM pre-training regime should focus mainly on skills and reasoning-based capabilities rather than knowledge, as advances in Retrieval Augmented Generation (RAG)\cite{asai2023selfrag, NEURIPS2020_6b493230, gao2024retrievalaugmented, lewis2020retrieval} improve feasibility of introducing updated knowledge at inference time. Besides being an effective way at combating hallucination\cite{shuster2021retrieval}, there is the added benefit of ensuring that the model's response will always be up-to-date without the need for continual learning. 

As such, we focused on collecting from data sources that contain information that is evergreen in nature, so as to maximize the proportion of ground truths in the pre-training dataset. For this reason, we avoided popular sources containing ephemeral content (e.g., subtitles, forums) or contemporary content (e.g., news). As part of the data collection effort, we also employed classifier models, text replacements, regex, and PDF cleaning tools to enhance quality.

\subsection{Selection of dataset via manual quality sampling}

With the current wide variety of language datasets, we could be more discriminatory in our dataset selection, thus significantly increasing the quality of our training corpus. First, we identified candidate datasets from published research, before randomly sampling from each dataset and scoring them. Based on a statistical approach, we reviewed 385 samples from each dataset to achieve 95\% confidence, and 5\% margin of error in our scoring (see Appendix~\ref{section:dataset_sel_proc}).

For construction of a "textbook/literature" corpus, we scored the pre-training samples on the 3 key attributes (see Appendix~\ref{section:dataset-judging-criteria}), using only "yes" and "no" for each attribute rather than a number scale to reduce subjectivity:

\begin{itemize}
    \item{Expository (2 points for a yes) - Whether the text explains or substantiates a concept, idea, or an opinion well.}
    \item{Toxic (-2 points for a yes) - Whether the text contains information that can be considered profanity, sexually inappropriate, racism, discrimination, extremism, or similar.}
    \item{Clean (1 point for a yes) - Whether the text contains irregular text sequences such as broken words, jumbled up text sequences, or garbled characters, and is generally free from excessive whitespace characters, irrelevant symbols, and any anomalies that may hinder the natural language processing.}
\end{itemize}

We then selected the highest-scoring datasets, until we met the target in terms of tokens counts.

For fine-tuning and alignment datasets, we observed that if the samples are synthetically generated, problematic samples tend to occur when it has sequence lengths in the 95\textsuperscript{th} percentile. Therefore, we reviewed all samples for the fine-tuning and alignment datasets in this region, and rejected the dataset if more than 10\% of the samples were found to exhibit hallucinations.

Despite our efforts, several limitations remained beyond our resourcing capabilities:
\begin{enumerate}
    \item{The factual correctness of samples, especially for complicated or lengthy ones, were too time-consuming to fully digest, and could not always be verified.}
    \item{With only two human judges, the margin of error impacted by human preferences would be higher. Ideally, more human judges could be used for the selection process.}
    \item{We are limited to evaluate samples in languages that the human judges are competent in, which is predominantly English.}
\end{enumerate}

In our future works, as the cost of compute reduces, alongside improvements in model capability, we plan to develop a pipeline for automated sampling and judging using AI. We would seek to avoid the use of proprietary AI, such as ChatGPT, for dataset selection to prevent any downstream free-use or legal issues.

\subsection{Selecting a model architecture}
We chose the Llama-2 architecture~\cite{touvron2023llama} for wider compatibility, with modification to the Llama-2 blocks by enlarging the Multi-layer Perceptron layers. To boost tokenization compression performance (by \textasciitilde4\%, see Appendix~\ref{appendix:mistral-tokenizer-vs-llama2}), we replaced Llama-2's tokenizer with a modified Mistral tokenizer. Our architectural choices are explained in section~\ref{section:model-architecture}.

\subsection{Fine-tuning and Alignment}
Aiming to produce a model predominantly useful as an AI assistant, we focused on maximizing its alignment with human preferences. Therefore, we manually curated a set of fine-tuning and reinforcement learning datasets to maximize the model's performance on MT-Bench. For alignment / reinforcement learning, we used Direct Preference Optimization (DPO)~\cite{rafailov2023direct}. Consistent with our overall dataset approach, the main attributes we looked for in fine-tuning and alignment datasets are expository explanations and well-elaborated answers. A breakdown of the fine-tune corpus is listed in Table \ref{table:finetuning-corpus}.

\section{Pre-Training Dataset}
\label{section:pre-training-dataset}
We constructed a pre-training dataset of 57 billion tokens, maintaining roughly 40\% for textbook/literature content, 40\% for web content, and 20\% coding content. This proportion mimicks the Phi-1.5 corpus~\cite{li2023textbooks}.

\begin{table}[h]
  \caption{Pre-Training Corpus}
  \label{table:pretraining-corpus}
  \centering
  \begin{tabular}{llll}
    \toprule
    Dataset & Number of Tokens & \% Tokens\\
    \midrule
    \\
    \underline{Textbook/literature}\\
    ArXiv & 9,859,118,710 & 17.31\\
    Wikipedia & 5,489,315,100 & 9.64\\
    US Public Domain Books & 4,096,982,180 & 7.19\\
    SciPhi/textbooks-are-all-you-need-lite & 558,437,415 & 0.98\\
    PhilArchive & 420,881,118 & 0.74\\
    Nampdn-ai/tiny-textbooks & 301,264,115 & 0.53\\
    Gutenberg & 288,233,832 & 0.51\\
    Nampdn-ai/tiny-orca-textbooks & 224,719,626 & 0.39\\
    Wikibooks & 213,767,786 & 0.38\\
    EuroParl Parallel Corpus & 74,046,955 & 0.13\\
    \midrule
    Subtotal & \textbf{21,526,766,837} & \textbf{37.80}\\
    \\
    \underline{Web content}\\
    Falcon-refinedweb & \textbf{22,814,264,174} & \textbf{40.07}\\
    \\
    \underline{Coding}\\
    Starcoder & \textbf{12,601,393,779} & \textbf{22.13}\\
    \midrule
    \midrule
    Total & \textbf{56,942,424,790} \\
    \bottomrule
  \end{tabular}
\end{table}

\subsection{Starcoder}
\label{section:starcoder}
The reasons for including a coding corpus are threefold. Firstly, introducing a coding corpus during the pre-training phase was shown to improve a model's reasoning capabilities~\cite{DBLP:journals/corr/abs-2211-09110, ma2024at}. Secondly, a language model that can code will have extended use-cases, such as for function/tool calling~\cite{schick2024toolformer}, or as a control agent~\cite{shen2024hugginggpt}. Thirdly, it supports a powerful "Program of Thought" methodology, which enhances problem-solving capabilities through generating code~\cite{DBLP:journals/corr/abs-2211-12588/programofthoughts}. The same observation is also reported in Phi-1.5's performance parity with models ~5x larger on the GSM8K math benchmark, by running the code generated in a virtual machine to obtain the final answer~\cite{li2023textbooks}. Table~\ref{table:pretraining-corpus} shows the specific breakdown of the pre-training dataset.

\subsection{ArXiv}
Arvix~\cite{arxiv_org_submitters_2024} not only provides for a high-quality collection of well-written and well-explained examples, it is also a rich source of knowledge and ground truths. We sub-sampled the ArXiv corpus in descending order (most recent entries first), until we reached an estimated 10 billion tokens. In addition, we converted all the LaTeX to Markdown format as it is a more general-purpose syntax.

\subsection{Wikipedia}
We took the English subset of Wikipedia~\cite{wikipedia}, and omitted articles with less than 1,000 characters as we find them to be of low quality.

\subsection{Books}
\textbf{US Public Domain Books}. This dataset was sub-sampled from \textit{storytracer/US-PD-Books}~\cite{huggingface2024uspdbooks}, which originally contained around 650,000 public domain books dating from 1521 to 1977. We obtain only the most recent set of books, by sub-sampling in descending date order, up to 4 billion tokens. This book corpus contains a mix of fiction and non-fiction material, which would provide the model with creativity and strong language capabilities. In addition, we noticed that the first few pages of each book contained content such as copyright notice, list of authors and content page, which would not add quality. Hence, for each book, we removed the first 200 lines, corresponding to an average of 3 pages.

\smallskip
\textbf{Gutenberg}. We filtered and included the English non-fiction books from Gutenberg~\cite{gutenberg} using the langdetect~\cite{langdetect} python package.

\smallskip
\textbf{Wikibooks}~\cite{dhruvil_dave_2021}. Similar to Gutenberg, we filtered out non-English books. Additionally, all HTML tags and "edit source" hyperlinks that exist within the dataset were also removed.

\subsection{PhilArchive}
PhilArchive~\cite{philarchive} is a non-profit project that is the largest open access e-print archive in philosophy. We included it in our dataset because philosophy is largely evergreen and provides high-quality reasoning and explanatory content.

\subsection{Euro Parliament proceedings}
We used the English subset of the Euro Parliament proceedings~\cite{koehn-2005-europarl} as a rich source of legal knowledge and well-substantiated debates on diverse perspectives.

\subsection{Synthetic data}
To obtain "textbook-like" synthetic data, we vetted numerous publicly available datasets, and avoided synthetic datasets that were generated by GPT3.5/4 due to licensing issues. The vetted dataset comprises of the following: \textit{SciPhi/textbooks-are-all-you-need-lite}~\cite{sciphi-textbooks}, \textit{Nampdn-ai/tiny-textbooks}~\cite{tiny-textbooks}, and \textit{Nampdn-ai/tiny-orca-textbooks}~\cite{tiny-orca-textbooks}.

\section{Model Architecture}
\label{section:model-architecture}

Tables ~\ref{table:model-architecture1} and \ref{table:model-architecture2} shows the model hyperparameters that we have adopted.

\begin{table}[H]
  \caption{Model Architecture (1)}
  \label{table:model-architecture1}
  \centering
  \begin{tabular}{llll}
    \toprule
    Parameters     & Vocab Size     & Embedding Size    & Context Length\\
    \midrule
    1,565,886,464 & 32,064  & 2048 & 16,384\\
    \bottomrule
  \end{tabular}
\end{table}

\begin{table}[H]
  \caption{Model Architecture (2)}
  \label{table:model-architecture2}
  \label{pretraining-hyperparams2}
  \centering
  \begin{tabular}{llll}
    \toprule
    Layers     & Heads     & Query Groups & Intermediate Hidden Size \\
    \midrule
    24 & 32 & 4 & 8192    \\
    \bottomrule
  \end{tabular}
\end{table}

We considered several state-of-the-art architectures, namely, Llama, GPTNeoX, and Mixtral Mixture-of-Experts. Eventually, we adopted the Llama-2 architecture for wider compatibility, with the following key modifications:

\subsection{Tokenizer}
Instead of using the Llama tokenizer, we adopted the Mistral tokenizer, with further modifications to improve its capabilities. The tokenizer is a critical component in the overall architecture as it sits upstream of the transformer architecture, where it converts input text sequences into numbers for downstream processing. As such, a powerful tokenizer will have a significant impact on model performance. 

\subsubsection{Rationale for Choosing Mistral Over Llama-2 Tokenizer}

Although the Llama-2 tokenizer remains the mainstream choice, we found the Mistral tokenizer to be superior in tokenization efficiency (higher compression), producing 3.61\% fewer tokens. The methodology and results are available in Appendix~\ref{appendix:mistral-tokenizer-vs-llama2}.

From the high-level perspective of data compression algorithms and language models as global approximation functions, a tokenizer that produces less tokens for any given text sequence directly improves the "bits per character (BPC)" metric, given by~\cite{huang2024compression}:
\[BPC = \frac{-\log_{2}{p_{model}(X)}}{T}=\frac{\displaystyle\sum\limits_{i=1}^n -\log_{2}{p_{model}(x_i\mid x_{1:i-1})}}{T}\]
where \(X\) is the corpus to be compressed, \(N\) is the total number of tokens of \(X\) tokenized by the model’s tokenizer and \(T\) is the number of characters of \(X\). 

A recent study established token compression as a strong indicator of "model intelligence"~\cite{huang2024compression} by Huang et al. As such, we opted for the Mistral tokenizer over Llama-2 to optimize the BPC for 1.5-Pints.

At the time of our pre-training, Llama-3 was yet to be released. Looking forward, we plan to explore the Llama-3 tokenizer in the same manner and report our findings.

\subsubsection{Pad Tokens}
Tokenizers from foundational models commonly lack the padding token, which is often necessary for many downstream use cases such as batch processing or model alignment, where sequences need to be padded to equal length. This results in the need to add the padding token retrospectively, which introduces three issues.

Firstly, it alters the vocabulary size and, consequently, the dimension of the language model head. This alteration requires additional coding logic to resize the weights (embedding layers) of the model head to the new vocabulary size.

Secondly, if the new vocabulary size is not divisible by 64, there could be a reduction in model throughout of up to 25\% ~\cite{karpathy24}. The vocabulary size could be arbitrarily extrapolated to the nearest multiple of 64, which again requires additional coding logic.

Thirdly, the absence of a padding token can lead to the common mistake of using the end-of-sequence token as a substitute, which provides an inaccurate representation for the model on when to accurately produce the end-of-sequence token to indicate that the generation should end. Another common workaround employed is the use of the unknown (\verb|<unk>|) token, which is also fundamentally incorrect.

Therefore, considering the near-universal necessity of a padding token and potential downstream logistical inconveniences and pitfalls, we decided to preemptively include the padding token (\verb`<|pad|>`) and extend the vocabulary size to 32,064 (from Mistral’s original 32,000). The model is pre-trained with this extended tokenizer from the start.

\subsubsection{Common chat template tokens}
As part of extending the vocabulary size to accommodate the padding token, we also added commonly-used chat template tokens. This makes the model versatile and ready for instruct fine-tuning out-of-the-box. Table~\ref{table:chat-template-tokens} shows the lists of chat templates tokens added our tokenizer.

\begin{table}[H]
  \caption{Chat Template Tokens}
  \label{table:chat-template-tokens}
  \centering
  \begin{tabular}{ll}
    \toprule
    Template & Tokens \\
    \midrule
    OpenAI ChatML & \verb+<|im_start|>+\\
    & \verb+<|im_end|>+\\
    \midrule
    Llama-2 & \verb+[INST]+\\
    & \verb+[/INST]+\\
    & \verb+<<SYS>>+\\
    & \verb+<</SYS>>+\\
    \midrule
    Llama-3 & \verb+<|begin_of_text|>+\\
    & \verb+<|start_header_id|>+\\
    & \verb+<|end_header_id|>+\\
    & \verb+<|eot_id|>+\\
    \midrule
    OpenChat & \verb+<|end_of_turn|>+\\
    \midrule
    Huggingface Zephyr & \verb+<|user|>+\\
    &\verb+<|system|>+\\
    &\verb+<|assistant|>+\\
    \bottomrule
  \end{tabular}
\end{table}

\subsubsection{Reserved token spaces for future customizability}
The tokenizer contains 49 remaining empty (\verb+<|reserved_n|>+) token spaces. These can be easily replaced with new tokens, which allows for ease of experimentation and fine-tuning on new chat templates.

\subsection{Sequence Length}
We explored 1.5-Pints with sequence length of both 2,048 and 16,384. As the latter represents a 2x increase over Llama-2 (8,192) and 8x increase over Apple OpenELM and Microsoft Phi (2,048), it allows for much wider application and downstream use cases.

\subsection{Grouped Query Attention}
We added Grouped Query Attention (not previously used in original smaller-sized Llama models), motivated by the need to improve speed in a bottlenecked autoregressive decoder model without noticeable quality degradation (as noted by Ainslie et. al.~\cite{ainslie2023gqa}). 

\subsection{Hidden Size}
Similar to Phi-1.5, we opted for a larger intermediate hidden size of 8,192, thereby increasing the dimensionality of the Multi-layer Perceptron (MLP). This is because recent studies have shown that increasing the dimensionality of MLPs improves their performance~\cite{Bachmann2023ScalingMA}.

\section{Pre-Training Approach}

\subsection{Hyperparameters}
Pre-train hyperparameters were referenced from TinyLlama~\cite{zhang2024tinyllama} and StableLM~\cite{bellagente2024stable}.

\begin{table}[H]
  \caption{Pre-training hyperparameters}
  \label{table:pretraining-hyperparameters}
  \centering
  \begin{tabular}{ll}
    \toprule
    Hyper-parameter & Value \\
    \midrule
    Optimizer & AdamW $(\beta_{1} = 0.9, \beta_{2} = 0.95)$ \\
    Learning Rate Scheduler & Cosine\\
    Max Learning Rate & $4.0\times10^{-4}$\\
    Min Learning Rate & $4.0\times10^{-5}$\\
    Warmup steps & 2,000 \\
    Batch size & 2,097,152 \\
    Weight Decay & 0.1 \\
    Gradient Clipping Threshold & 1.0 \\
    \bottomrule
  \end{tabular}
\end{table}

\subsection{Training duration}

We pre-trained 1.5-Pints following a standard autoregressive sequence modeling for a total of 2 epochs, trained on 8 x A100s. Table~\ref{table:training-duration} shows the training duration breakdown for all phases of the model training. Table~\ref{table:hardware-performance} shows the hardware performance.

\begin{table}[H]
  \caption{Training duration}
  \label{table:training-duration}
  \centering
  \begin{tabular}{lll}
    \toprule
    Stage & Epochs & Duration \\
    \midrule
    Pre-training & 2 & 8d 2h\\
    Fine-tuning & 5 & 21h\\
    DPO & 5 & 3h\\
    \midrule
    & Total & 9d 2h\\
    \bottomrule
  \end{tabular}
\end{table} 

\begin{table}[H]
  \caption{Hardware performance for pre-training.}
  \label{table:hardware-performance}
  \centering
  \begin{tabular}{llll}
    \toprule
    GPU type & Throughput (TFLOPS/GPU/s) & Tokens/GPU/s & Utilisation (\%) \\
    \midrule
    A100 & 199.97 & 17,528 & 99.61\\
    \bottomrule
  \end{tabular}
\end{table} 

\section{Fine-tuning Approach}
Aligning with the main research aim of 1.5-Pints, we approached fine-tuning with a particular focus on data, paying particular attention to selecting datasets based on the quality of the responses. In particular, we sought clear, well-explained and sufficiently elaborated answers. To increase the diversity of instruction-following examples, a wide variety of fine-tuning datasets (as illustrated in Table~\ref{table:finetuning-corpus}) was included. 

\begin{table}[H]
  \caption{Fine-Tuning Corpus}
  \label{table:finetuning-corpus}
  \centering
  \begin{tabular}{llll}
    \toprule
    Dataset & Number of Tokens & \% Tokens\\
    \midrule
    HuggingFaceH4/ultrachat\_200k & 270,940,328 & 37.35\\
    Open-Orca/SlimOrca-Dedup & 147,987,500 & 20.40\\
    meta-math/MetaMathQA & 98,590,031 & 13.59\\
    HuggingFaceH4/deita-10k-v0-sft & 86,003,252 & 11.86\\
    WizardLM/WizardLM\_evol\_instruct\_V2\_196k & 79,142,695 & 10.91\\
    togethercomputer/llama-instruct & 25,865,812 & 3.57\\
    LDJnr/Capybara & 16,842,409 & 2.32\\
    \midrule
    \midrule
    Total & 725,372,027\\
    \bottomrule
  \end{tabular}
\end{table}

Of note, we included the \textit{Open-Orca/SlimOrca-Dedup} as the dataset is premised upon imbuing smaller models with the reasoning capabilities of a large model through a "step-by-step" reasoning. Specifically, as inspired by Chain of Thought reasoning~\cite{wei2023chainofthought}, the \textit{Open-Orca/SlimOrca-Dedup} dataset focuses on making models generate answers generated "step-by-step" explanation traces, before finally arriving at the final answer~\cite{mukherjee2023orca}. We pre-processed \textit{Open-Orca/SlimOrca-Dedup} by training a small BERT model with classification heads to filter poor samples.

Additionally, we included \textit{HuggingFaceH4/ultrachat\_200k} and \textit{WizardLM/WizardLM\_evol\_instruct\_V2\_196k} (both of which were generated using the \textit{Evolve-Instruct} method) as such datasets increases the complexity of the instructions, which had been proven to help fine-tuned models follow complex instructions well~\cite{xu2023wizardlm}. 

We fine-tuned for a total of 5 epochs using the hyperparameters (referenced from Zephyr Alignment Handbook~\cite{alignment_handbook2023}) listed in Table \ref{table:finetuning-hyperparameters}, before selecting epoch 2 as the best epoch for Reinforcement Learning step.

\begin{table}[H]
  \caption{Fine-tuning hyperparameters}
  \label{table:finetuning-hyperparameters}
  \centering
  \begin{tabular}{ll}
    \toprule
    Hyper-parameter & Value \\
    \midrule
    Optimizer & AdamW $(\beta_{1} = 0.9, \beta_{2} = 0.95)$ \\
    Batch size & 1,048,512 \\
    Warmup steps & 1,126 (10\%) \\
    Peak learning rate & 2e-5 \\
    Learning rate scheduler & Cosine \\
    Weight Decay & 0.1 \\   
    \bottomrule
  \end{tabular}
\end{table}

\section{Reinforcement Learning Approach}
For our alignment step via reinforcement learning, we opted for Direct Preference Optimization (DPO)~\cite{rafailov2023direct} using the Ultrafeedback dataset~\cite{cui2023ultrafeedback}, where we followed the approach described in The Alignment Handbook~\cite{alignment_handbook2023}. We ran for a total of 5 epochs, before selecting epoch 2 as the best epoch (see Appendix~\ref{appendix:epoch-study-dpo}).

\section{Results}
We opted to evaluate 1.5-Pints on MT-Bench, as it represents the closest an automated evaluation benchmark can approximate human evaluation, traditionally considered the gold standard in natural language processing. MT-Bench achieves this by using GPT-4 to rate responses on a scale of 1 to 10. We ran the evaluation for all models with repetition penalties of 1.0 and 1.3, and selected the higher of the two as the model's representative score (refer to Table~\ref{table:MT-Bench-results} for the breakdown of results).

\begin{table}[H]
  \caption{MT-Bench Comparison with SOTA models under 3B (Llama-2-7b is included for reference)}
  \label{table:MT-Bench-results}
  \centering
  \begin{tabular}{>{\raggedright\arraybackslash}p{5.5cm}>{\centering\arraybackslash}p{0.75cm}>{\centering\arraybackslash}p{0.75cm}>{\centering\arraybackslash}p{1.5cm}>{\centering\arraybackslash}p{1.5cm}>{\centering\arraybackslash}p{1.5cm}}
    \toprule
    \multirow{2}{*}{Model} & \multicolumn{2}{c}{\shortstack{Repetition\\Penalty}} & \multirow{2}{*}{\shortstack{Parameter\\Size}} & \multirow{2}{*}{\shortstack{Pre-train\\Tokens}} & \multirow{2}{*}{\shortstack{Context\\Window}}\\
    & 1.3 & 1.0 \\
    \midrule
    meta-llama/Llama-2-7b-chat-hf & - & \textbf{6.27} & 7B & 2T & 4K\\
    \midrule
    microsoft/phi-2 & 3.83 & \textbf{5.83} & 2.7B & 1.4T & 2K\\
    google/gemma-2b-it & 5.31 & \textbf{5.44} & 2B & 3T & 8K\\
    stabilityai/stablelm-2-1\_6b-chat & 2.88 & \textbf{4.7} & 1.6B & 2T & 4K\\
    \rowcolor{lightergray} 1.5-Pints-2K & \textbf{3.73} & 2.52 & 1.57B & \textbf{0.115T} & 2K\\
    TinyLlama/TinyLlama-1.1B-Chat-v1.0 & \textbf{3.72} & 3.53 & 1.1B & 3T & 2K\\
    \rowcolor{lightergray} 1.5-Pints-16K & \textbf{3.40} & 2.52 & 1.57B & \textbf{0.115T} & \textbf{16K}\\
    apple/OpenELM-1\_1B-Instruct & \textbf{3.34} & 2.26 & 1B & 1.8T & 2K\\
    microsoft/phi-1\_5 & 1.13 & \textbf{3.33} & 1.3B & 0.15T & 2K\\
    databricks/dolly-v2-3b & \textbf{2.33} & 1.46 & 3B & 0.3T & 2K\\
    EleutherAI/pythia-2.8b & \textbf{1.81} & 1.48 & 2.8B & 0.3T & 2K\\
    tiiuae/falcon-rw-1b & \textbf{1.18} & 1.08 & 1B & 0.35T & 2K\\
    \bottomrule
  \end{tabular}
\end{table}

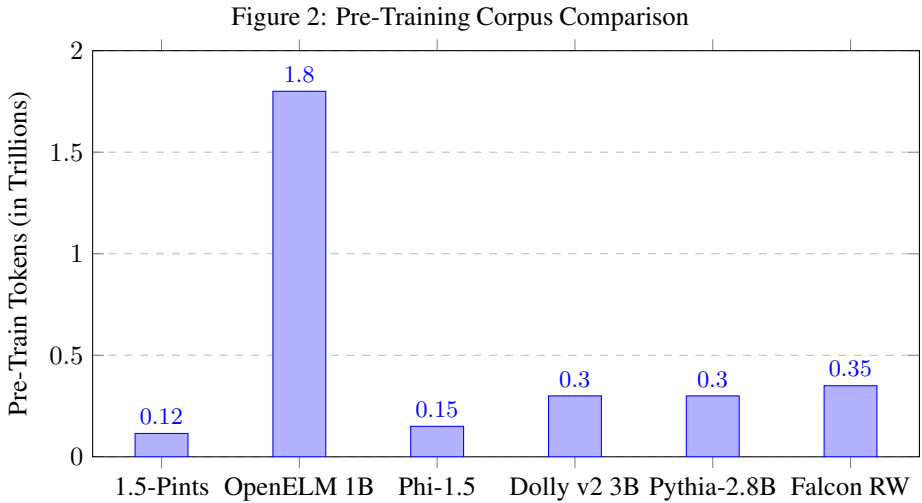
\begin{figure}[h]
    \centering
    \caption{Pre-Training Corpus Comparison}
    \label{figure:comparison-pretraining-corpus}
    \begin{tikzpicture}
    \begin{axis}[
        ybar,
        width=0.9\textwidth,
        height=0.5\textwidth,
        symbolic x coords={1.5-Pints, OpenELM 1B, Phi-1.5, Dolly v2 3B, Pythia-2.8B, Falcon RW},
        xtick=data,
        ylabel={Pre-Train Tokens (in Trillions)},
        ytick={0, 0.5, 1.0, 1.5, 2.0},
        ymin=0, ymax=2.0,
        ymajorgrids=true,
        grid style=dashed,
        bar width=20pt,
        nodes near coords,
        nodes near coords align={vertical},
        every node near coord/.append style={font=\small},
    ]
    
    \addplot
        coordinates {
            (1.5-Pints,0.115)(OpenELM 1B,1.8)(Phi-1.5,0.15)(Dolly v2 3B,0.3)(Pythia-2.8B,0.3)(Falcon RW, 0.35)
        }; 
    \end{axis}
    \end{tikzpicture}
\end{figure}

 Under this benchmark, 1.5-Pints of both 2k and 16k context sizes demonstrated better performance compared to popular models such as OpenELM 1.1B Instruct, Pythia 2.8B, Phi 1.5, and Falcon RefinedWeb 1B, despite being pre-trained on a much smaller amount of tokens (see Figure \ref{figure:comparison-pretraining-corpus}).

While traditional benchmarks are not our focus, we have also evaluated our model on the standard tasks found in Huggingface OpenLLM leaderboard using the EleutherAI Evaluation Harness (see Appendix~\ref{appendix:language-model-evaluation-harness-summary}), where our model performed comparably with other SOTA models despite having the lowest number of pre-training tokens on the list.

\section{Future Developments}
Given that there is a finite amount of textbook or expository grade material available for model pre-training, synthetic corpus generation becomes an essential component of success ~\cite{gunasekar2023textbooks, mukherjee2023orca, xu2023wizardlm}, especially for niche domains where corpora is limited. While this is already possible with well-crafted prompting~\cite{lee2023making}, such models can still hallucinate in niche knowledge domains~\cite{bouyamourn2023why}. Concomitantly, increasing the accuracy of synthetically generated datasets remains an important area of research. Thus, we have identified several areas of development to push for more efficient, accurate, and scalable corpus generation methodologies.

\subsection{Retrieval Augmented Generation (RAG)}
RAG has been popularized as an effective method in generating expository-grade texts using a set of ground-truth corpora. Traditionally, RAG pipelines for LLMs hinge mainly on lookup within a vector database of chunked document vectors using a similarity metric for ensuring relevance~\cite{gao2024retrievalaugmented}. Subsequently, RAG has also gained traction for its ability to reduce hallucination by grounding the model in a corpora of domain-specific "truths"~\cite{shi2023replug, shuster2021retrieval, shi-etal-2022-xricl}, with recent works seeking to enhance the quality of such retrieval~\cite{asai2023selfrag, kang2024ever}. An alternative paradigm of RAG exists in architectural integration of retrieval with foundational LLM models~\cite{NEURIPS2020_6b493230}. However, much work remains to be done in the realm of RAG-pipeline and RAG-model optimization, to ensure that such a solution remains feasible for generating datasets en-masse, with some interesting pioneering works being pluggable virtual tokens~\cite{zhu2024token} and binary token representations~\cite{cao2024btr}). 

\subsection{Knowledge Graphs}
Building on the success of the RAG-based method, Knowledge Graph prompt-augmentation has shown potential for limiting hallucination in LLM generation~\cite{soman2024biomedical, wang2023knowledge}. This method works by injecting knowledge from the Knowledge Graph into prompts sent to the LLM. By constraining an LLM to the data stored within a Knowledge Graph, the generator-LLM is forced to adhere to vetted "knowledge"\cite{sui2024fidelis}, rather than allowing the generator-LLM to rely solely on next-token prediction which is vulnerable to hallucination due to exposure bias\cite{10.1145/3571730}. Improving and leveraging Knowledge Graphs could further reduce hallucination risk in synthetic data.

\subsection{Tool-based retrieval}
An alternative pathway to Knowledge Graph based methods that takes inspiration from RAG would be tool-based methods, such as GPT4Tools~\cite{NEURIPS2023_e3936777} or ToolkenGPT~\cite{hao2023toolkengpt}. These methods leverage external tools that a model fine-tuned for this purpose has at its disposal, allowing it to augment its pre-trained knowledge with data that it can harness from these tools. This paradigm represents an exciting area of research due to the increased possibility of having a hallucination-free high-quality synthetically-generated dataset, even as the LLM scene pushes towards LLM-agents~\cite{prakash2023llm, NEURIPS2023_a3621ee9, kim2023prospector}.

\section{Conclusion}
In this technical report, we presented "1.5 Pints", a Large Language Model that significantly advances the efficiency of LLM training by emphasizing data quality over quantity. Our model was trained on a meticulously curated corpus of 57 billion tokens, thereby demonstrating that a smaller, high-quality dataset can surpass the performance of models trained on much larger datasets. This approach not only reduced the training time precipitously (from months to just days), but also minimized the computational resources required, thus making pre-training more accessible and environmentally-friendly.

\subsection{Key Performance Attribution}
We primarily attribute the performance of 1.5-Pints to the following:
\begin{enumerate}
\item Careful selection of high-quality data forms a solid foundation for the LLM to perform well despite being trained on much lesser data.
\item A meticulously crafted model architecture that leverages the best of SOTA practices distilled from Mistral, StableLM, and TinyLlama.
\end{enumerate}

\subsection{Open-Source}
By open-sourcing our findings and resources, we make our experiment verifiable and also aim to support the research community in the development of more efficient and powerful LLMs. This work underscores the critical role of data quality in LLM training and sets a new benchmark for resource-efficient AI development. Moving forward, we anticipate that broader adoption of these principles will drive innovation and enhance accessibility in the field of artificial intelligence.

\section{Acknowledgements}
We would like to acknowledge the following people for their invaluable assistance to our experiment:
\begin{itemize}
    \item The StatNLP Research Group and the TinyLlama Team from the Singapore University of Technology and Design for their significant contributions in the arena of small language models developments.
    \item Kang Pyo, Lee Zhan Peng and Kay Eugenia Purnama for assisting in bug fixing and benchmarking.
    \item Nolan Nguyen and Annie Nguyen from Google Cloud Platform for enrolling Pints.ai into the GenAI startup program, which provided GPU credits for our experiments.
    \item The HuggingFace Community for freely sharing their datasets and models.
\end{itemize}

\pagebreak
\bibliographystyle{plain} 
\bibliography{main} 

\pagebreak
\section{Appendix}
\begin{appendices}

\section{Manual Dataset Selection Procedure via Central Limit Theorem and Population Sampling}
\label{section:dataset_sel_proc}
To ensure that we select the best datasets for pre-training, we sought to manually inspect the examples and select the dataset which scored the highest. However, with a huge corpus size, it is not feasible to manually inspect the entire dataset. Thus, we sampled 385 examples from the corpora, the number of which was derived by the following Population Sampling formula:

\[ n = \lceil \frac{Z^2\times p\times (1-p)}{e^2} \rceil \]

where \(n\) is the least sample size, \(Z\)  is the Z-value (which we used 1.96 for a 95\% confidence level), \(p\) is the estimated proportion of outcomes, which we used 50\% as a default for unknown true proportions, and \(e\) is the margin of error, which we used 5\%, resulting in:

\[ n = \lceil \frac{1.96^2\times 0.5\times (1-0.5)}{0.05^2} \rceil = 385\] 

\section{Manual Dataset Selection Judging Criteria}
\label{section:dataset-judging-criteria}

For scoring the pre-training dataset, two human judges are employed to judge the the datasets based on the following criteria, and only scoring on the basis of yes or no to reduce subjectivity:

\begin{table}[H]
  \caption{Judging Criteria}
  \label{table:judging-criteria}
  \centering
  \setlength{\leftmargini}{0.4cm}
  \begin{tabular}{ m{2cm} m{1.5cm} m{10cm} }
    \toprule
    Attribute & Points & Criteria\\
    \midrule
    Expository & 2 & Whether the text explains or substantiates a concept, idea, or an opinion well.\newline
    \begin{itemize} 
        \item Does the text provide clear and detailed explanations of concepts or ideas? 
        \item Does the text include examples or evidence to support the main points?
        \item Is the information presented in a logical and organized manner?
        \item Does the text define any specialized terms or jargon used?
        \item Is the purpose of the text to inform, explain, or clarify rather than to persuade or entertain?
    \end{itemize} \\
    \midrule
    Toxic & -2 & Whether the text contains information that can be considered profanity, sexually inappropriate, racism, discrimination, extremism, or similar.\newline
    \begin{itemize} 
        \item If you read this text, would you find any language or words offensive? 
        \item Does this text make you uncomfortable due to sexually inappropriate content?
        \item If this text was directed at you or your group, would you feel discriminated against or demeaned?
        \item Does the text express or promote harmful or extremist ideas that you find alarming?
        \item Do you think this text would be hurtful or harmful to others based on their race, gender, sexual orientation, religion, or ethnicity? 
    \end{itemize} \\
    \midrule
    Clean & 1 & Whether the text contains irregular text sequences.\newline
    \begin{itemize}
        \item Are there a too many broken words (more than 1 in 10 words)?
        \item Are there jumbled up text sequences (more than 1 in 5 paragraphs)?
        \item Are there irrelevant symbols (more than 1 in 5 paragraphs)?
        \item Are there excessive whitespaces (more than 1 in 10 words)?
    \end{itemize}\\
    \bottomrule
  \end{tabular}
\end{table}

For scoring the fine-tuning and alignment dataset, the "clean" attribute is omitted, as the handcrafted or synthetically generated samples are clean to being with.

\section{Mistral Tokenizer vs Llama-2 Tokenizer}
\label{appendix:mistral-tokenizer-vs-llama2}
To investigate the tokenization efficiency between the Mistral tokenizer and the Llama-2 tokenizer, we randomly subsampled 10\% of the \verb|tiiuae/falcon-refinedweb|\cite{technology_innovation_institute_2023} dataset. Of note, both the Mistral and the Llama-2 tokenizers have the same vocabulary size of 32K, which makes this comparison a fair one.

\begin{table}[H]
  \caption{Tokenization}
  \label{table:tokenization}
  \centering
  \begin{tabular}{lll}
    \toprule
    Tokenizer & Tokens & \% Reduction\\
    \midrule
    Llama-2 & 24,131,968,012 & -\\
    Mistral & 23,261,356,142 & 3.61\\
    \bottomrule
  \end{tabular}
\end{table}

We also found that the metadata within the Mistral \verb|tokenizer.model| indicates it was trained with a corpus named \verb|@/mnt/test/datasets/tokenizer_training/8T_train_data/shuffled.txt|, suggesting that the Mistral tokenizer was trained using approximately 8 trillion words or tokens. This contrasts with the Llama-2 tokenizer, which is derived from its predecessor, Llama-1~\cite{touvron2023llama}, and was likely trained on a significantly smaller corpus. This substantial difference in training data volume could account for the performance improvements observed in the Mistral tokenizer over Llama-2.

\section{Direct Preference Optimization Epoch Study}
\label{appendix:epoch-study-dpo}
We conducted 5 epochs of DPO training and evaluated all epochs on MT-Bench. For 2K context window, the best MT-Bench score was achieved by epoch 2 of the DPO step. We noted that further epochs beyond epoch 2 provided no improvement, and with a significant dip in epoch 4. For 16K context window, the epoch 4 achieved the best result, before dipping in performance at epoch 5. We used a repetition penalty of 1.3 for the evaluations.

\begin{table}[H]
  \caption{Epoch study for 2K Context Window DPO with MT-Bench}
  \label{dpo-epoch-2k}
  \centering
  \begin{tabular}{lll}
    \toprule
    Epoch & Checkpoint Steps & MT-Bench Score \\
    \midrule
    1 & 1,910 & 3.67 \\
    \rowcolor{lightergray} 2 & 3,821 & 3.73 \\ 
    3 & 5,731 & 3.62 \\
    4 & 7,642 & 3.43 \\
    5 & 9,550 & 3.61 \\
    \bottomrule
  \end{tabular}
\end{table}

\label{appendix:epoch-study-dpo-16K}
\begin{table}[H]
  \caption{Epoch study for 16K Context Window DPO with MT-Bench}
  \label{dpo-epoch-16k}
  \centering
  \begin{tabular}{lll}
    \toprule
    Epoch & Checkpoint Steps & MT-Bench Score \\
    \midrule
    1 & 1,910 & 3.12 \\
    2 & 3,821 & 3.10 \\ 
    3 & 5,731 & 3.16 \\
    \rowcolor{lightergray} 4 & 7,642 & 3.40 \\
    5 & 9,550 & 3.30 \\
    \bottomrule
  \end{tabular}
\end{table}

\section{Language Model Evaluation Harness}
\label{appendix:language-model-evaluation-harness-summary}

Similar to 1.5-Pints' performance on MT-Bench, it also performed well on LM-Eval, out-performing models such as EleutherAI's GPTNeo 1.3B model, as well as larger models such as StabilityAI's StableLM 3b fine-tuned model. 

\begin{table}[H]
  \caption{Language Model Evaluation Results (Summary)}
  \label{lmeval-harness-results}
  \centering
  \begin{tabular}{p{5.5cm}p{1.5cm}p{2.5cm}p{1.5cm}}
    \toprule
    Model & Average & Pre-train Tokens & Parameter Size \\
    \midrule
    microsoft/phi-2 & 61.09 & 1.4T & 2.7B\\
    TinyLlama/TinyLlama-1.1B-Chat-v1.0 & 37.28 & 3T & 1.1B\\
    EleutherAI/pythia-1.4b-deduped & 35.00 & 0.3T & 1.4B\\
    facebook/opt-1.3b & 34.60 & 0.18T & 1.3B\\
    \rowcolor{lightergray}1.5-Pints-16K & 34.06 & 0.115T & 1.57B\\
    \rowcolor{lightergray}1.5-Pints-2K & 34.02 & 0.115T & 1.57B\\
    EleutherAI/gpt-neo-1.3B & 33.59 & 0.38T & 1.3B\\
    bigscience/bloom-1b & 32.48 & 0.35T & 1B\\
    stabilityai/stablelm-tuned-alpha-3b & 32.14 & 2T & 3B\\
    stabilityai/stablelm-base-alpha-3b & 31.50 & 2T & 3B\\
    facebook/xglm-1.7B & 31.42 & 0.5T & 1.7B\\
    bigcode/starcoderbase-3b & 31.37 & 1T & 3B\\
    cerebras/Cerebras-GPT-1.3B & 31.30 & 0.371T & 1.3B\\
    \midrule
    apple/OpenELM-1\_1B-Instruct  & 49.94* & 1.8T & 1.1B\\
    \bottomrule
  \end{tabular}
\end{table}

*Although Apple OpenELM did not report GSM8K results, we included the average of its known metrics at the bottom of the table for reference.

\begin{table}[H]
  \caption{Language Model Evaluation Results (Detailed)}
  \label{lmeval-harness-results-detailed}
  \centering
  \begin{tabular}{p{3.9cm}p{0.75cm}p{1.25cm}p{0.8cm}p{1.4cm}p{1.5cm}p{1.25cm}}
    \toprule
    Model & ARC & HellaSwag	& MMLU & TruthfulQA	& Winogrande & GSM8K \\
    \midrule
    phi-2 & 61.01 & 74.92 & 57.92 & 44.24 & 73.48 & 54.97 \\
    TinyLlama-1.1B-Chat-v1.0 & 36.09 & 61.1 & 25.39 & 37.48 & 61.25 & 2.35 \\
    pythia-1.4b-deduped & 32.68 & 54.96 & 25.56 & 38.66 & 57.3 & 0.83 \\
    opt-1.3b & 29.52 & 54.53 & 24.96 & 38.71 & 59.75 & 0.15 \\
    \rowcolor{lightergray}1.5-Pints-16K & 29.01 & 45.39 & 26.52 & 38.93 & 53.04 & 11.45\\
    \rowcolor{lightergray}1.5-Pints-2K & 30.55 & 48.43 & 25.39 & 40.13 & 53.51 & 6.14\\
    gpt-neo-1.3B & 31.23 & 48.47 & 24.82 & 39.63 & 56.91 & 0.45 \\
    bloom-1b & 28.33 & 42.78 & 26.7 & 41.8 & 55.01 & 0.23 \\
    stablelm-tuned-alpha-3b & 27.82 & 44.06 & 23.08 & 42.33 & 55.01 & 0.53 \\
    stablelm-base-alpha-3b & 26.45 &	42.24 &	25.43 &	40.5 &	53.91 &	0.45 \\
    facebook xglm-1.7B & 25.85 & 45.68 & 25.1 &	37.21 &	53.91 &	0.76 \\
    starcoderbase-3b &	25.85 &	39.11 &	27.35 &	43.05 &	51.14 &	1.74\\
    Cerebras-GPT-1.3B &	26.28 &	38.54 &	26.59 &	42.7 &	53.43 &	0.23\\
    \midrule
    OpenELM-1\_1B-Instruct  & 41.55 & 71.83 & 25.65 & 45.95 & 64.72 & - \\
    \bottomrule
  \end{tabular}
\end{table}

\section{Tracking the effects of each stage of training}
The effects of each stage of training is summarized in Figure~\ref{figure:modality-impact}:

\begin{figure}[h]
    \centering
    \caption{Impact of training modality on performance}
    \label{figure:modality-impact}
    \begin{tikzpicture}
    \begin{axis}[
        ybar,
        xlabel={Stage},
        symbolic x coords={Pre-train, SFT, DPO},
        xtick=data,
        ylabel={MT-Bench Score},
        ytick={0,1,2,3,4},
        ymin=0, ymax=4,
        ymajorgrids=true,
        grid style=dashed,
        enlarge x limits=0.3,
        bar width=20pt,
        legend style={at={(0.5,-0.1)},
	    anchor=north,legend columns=-1},
    ]
    \addplot
        coordinates {
            (Pre-train,1)(SFT,3.01)(DPO,3.73)
        }; 
    \addplot
        coordinates {
            (Pre-train,1.05)(SFT,2.68)(DPO,3.40)
        };
    \legend{1.5-Pints-2K,1.5-Pints-16K}
    \end{axis}
    \end{tikzpicture}
\end{figure}
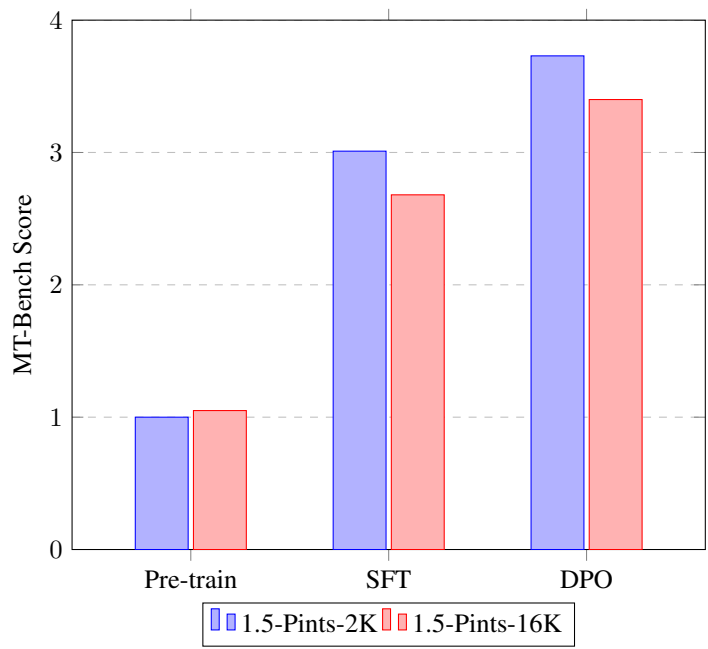

\section{Legal Warning}
Our model and training code are open-sourced under MIT license.

Though best efforts has been made to ensure, as much as possible, that all texts in the training corpora are royalty free, this does not constitute a legal guarantee that such is the case. \textbf{By using any of the models, corpora or part thereof, the user agrees to bear full responsibility to do the necessary due diligence to ensure that he/she is in compliance with their local copyright laws}. 

In addition, the \textbf{user agrees to bear any damages} arising as a direct cause (or otherwise) of using any artifacts released by the pints research team, and full responsibility for the consequences of his/her usage (or implementation) of any such released artifacts. The user also indemnifies Pints Research Team (and any of its members or agents) of any damage, related or unrelated, to the release or subsequent usage of any findings, artifacts or code by the team.

For the avoidance of doubt, \textbf{any artifacts released by the Pints Research team are done so in accordance with the "fair use"} clause of Copyright Law, in hopes that this will aid the research community in bringing LLMs to the next frontier.

\end{appendices}

\end{document}